\documentclass[11pt,a4paper]{article}
\usepackage[hyperref]{emnlp2018}
\usepackage{times}
\usepackage{latexsym}
\usepackage{graphicx}
\usepackage{booktabs}

\usepackage{url}

\usepackage[linesnumbered]{algorithm2e}

\aclfinalcopy 

\newcommand\mq{Macquarie University}

\title{\mq\ at BioASQ 6b: Deep learning and deep reinforcement learning for query-based multi-document summarisation}

\author{Diego Moll\'a \\
  Macquarie University \\
  Sydney, Australia \\
  {\tt diego.molla-aliod@mq.edu.au}} 

\date{}

\begin{document}
\maketitle
\begin{abstract}
This paper describes \mq's contribution to the BioASQ Challenge (BioASQ 6b, Phase B). We focused on the extraction of the ideal answers, and the task was approached as an instance of query-based multi-document summarisation. In particular, this paper focuses on the experiments related to the deep learning and reinforcement learning approaches used in the submitted runs. The best run used a deep learning model under a regression-based framework. The deep learning architecture used features derived from the output of LSTM chains on word embeddings, plus features based on similarity with the query, and sentence position. The reinforcement learning approach was a proof-of-concept prototype that trained a global policy using REINFORCE. The global policy was implemented as a neural network that used $tf.idf$ features encoding the candidate sentence, question, and context.
\end{abstract}

\section{Introduction}

The BioASQ Challenge\footnote{\url{http://www.bioasq.org/}} consists of various tasks related to biomedical semantic indexing and question answering \cite{Tsatsaronis:2015}. Our participation in BioASQ for 2018 focused on Task~B Phase~B, where our system attempted to find the ideal answer given a question and a collection of relevant snippets of text. We approached this task as an instance of query-based multi-document summarisation, where the ideal answer is the summary to produce.

The BioASQ challenge focuses on a restricted domain, namely biomedical literature. Nevertheless, the techniques developed for our system were domain-agnostic and can be applied to any domain, provided that the domain has enough training data and a specialised corpus large enough to train word embeddings.

We were interested in exploring the use of deep learning and reinforcement learning for this task. Thus, Section~\ref{sec:deep} explains our experiments using deep learning techniques. Section~\ref{sec:rl} details our experiments using reinforcement learning. Section~\ref{sec:settings} specifies the settings used in the experiments. Section~\ref{sec:results} shows and discusses the results, and Section~\ref{sec:conclusions} concludes the paper.

\section{Deep Learning}\label{sec:deep}

The deep learning experiments followed the general framework introduced by \citet{Molla2017}, which can be summarised as a regression approach that follows these three main steps:

\begin{enumerate}
\item Split the input text into candidate sentences.
\item Score each candidate sentence independently.
\item Return the $n$ sentences that have the highest score.
\end{enumerate}

In all of the experiments reported in this paper, the input text is the set of relevant snippets that are associated with the question. These snippets are pre-processed by splitting them into sentences. The sentences are then scored using the deep learning approaches described below. Then, after all candidate sentences are scored, the top $n$ sentences are returned as the ideal answer. The value of $n$ is determined empirically and it depends on the question type as shown in Table~\ref{tab:n}. These are the same settings as in \citet{Molla2017}'s framework.
\begin{table}
  \centering
  \begin{tabular}{ccccc}
   
    & \textbf{Summary} & \textbf{Factoid} & \textbf{Yesno} & \textbf{List}\\
    \midrule
    \textbf{n} & 6 & 2 & 2 & 3
  \end{tabular}
  \caption{Value of $n$ (the number of sentences returned as the
    ideal answer) for each question type.}\label{tab:n}
\end{table}

The deep learning experiments predict the score of an input sentence by applying supervised regression techniques. Following \citet{Molla2017}'s framework, the training data was annotated with the F1 ROUGE-SU4 score of each individual sentence using the ideal answer as the target. Also following \citet{Molla2017}'s framework, the architecture of our system architecture was based on the following main stages:

\begin{enumerate}
\item Obtain the word embedding of every word in the input sentence and the question. 
\item Given the word embeddings of the input sentence and the question, obtain the sentence and question embeddings.
\item Feed the sentence embeddings and the similarity between the sentence and question embeddings to a fully connected layer and final linear combination. In this stage, as an extension to \citet{Molla2017}'s approach, we also incorporated information about the sentence position.
\end{enumerate}

The word embeddings were obtained by pre-training word2vec \cite{Mikolov:2013} on a collection of PubMed documents made available by the organisers of BioASQ. Given a sentence (or question) $i$ with $n_i$ words and vectors of word embeddings $m_1$ to $m_{n_i}$, we ran experiments using the following two alternative approaches to obtain the sentence (or question) vector of embeddings $s_i$:

\begin{description}
\item[NNR Mean.] Compute the mean of the word embeddings:
$$
s_i = \frac{1}{n_i}\sum_{j=1}^{n_i}m_j
$$
\item[NNR LSTM.] Feed the sequence of word embeddings to bidirectional Recurrent Neural Networks with LSTM cells. We used TensorFlow's LSTM implementation, which is reportedly based on \citet{Hochreiter:1997}'s seminal work. More explicitly, the embedding $s_i$ of sentence $i$ is the concatenation of the output of the last cell in the forward chain ($\overrightarrow{h}_{n_i}$) and the first cell in the backward chain ($\overleftarrow{h}_1$):
$$
s_i  = \left[\overrightarrow{h}_{n_i};\overleftarrow{h}_1\right]
$$
\noindent Cell at position $t$ of the forward chain receives its input from the embedding of word at position $t$ and cell at position $t-1$:
$$
\begin{array}{rcl}
\overrightarrow{c}_t & = & \overrightarrow{f} \odot \overrightarrow{c}_{t+1} + \overrightarrow{i} \odot \overrightarrow{z}\\
\overrightarrow{h}_t & = & \overrightarrow{o} \odot \tanh(\overrightarrow{c}_t)\\
\overrightarrow{i} & = & \sigma(\overrightarrow{W}_{i}\cdot m_t + \overrightarrow{U}_{i}\cdot \overrightarrow{h}_{t+1} + \overrightarrow{b}_i)\\
\overrightarrow{f} & = & \sigma(\overrightarrow{W}_{f}\cdot m_t + \overrightarrow{U}_{f}\cdot \overrightarrow{h}_{t+1} + \overrightarrow{b}_f)\\
\overrightarrow{o} & = & \sigma(\overrightarrow{W}_{o}\cdot m_t + \overrightarrow{U}_{o}\cdot \overrightarrow{h}_{t+1} + \overrightarrow{b}_o)\\
\overrightarrow{z} & = & \tanh(\overrightarrow{W}_{z}\cdot m_t + \overrightarrow{U}_{z}\cdot \overrightarrow{h}_{t+1} + \overrightarrow{b}_z)\\
\end{array}
$$
\noindent where $\sigma$ is the logistic sigmoid function, $\odot$ is the element-wise multiplication of two vectors, and $\cdot$ is the dot product between a matrix and a vector. 

Cell at position $t$ in the backward chain receives its input from $m_t$ and cell at position $t+1$:
$$
\begin{array}{rcl}
\overleftarrow{c}_t & = & \overleftarrow{f} \odot \overleftarrow{c}_{t+1} + \overleftarrow{i} \odot \overleftarrow{z}\\
\overleftarrow{h}_t & = & \overleftarrow{o} \odot \tanh(\overleftarrow{c}_t)\\
\overleftarrow{i} & = & \sigma(\overleftarrow{W}_{i}\cdot m_t + \overleftarrow{U}_{i}\cdot \overleftarrow{h}_{t+1} + \overleftarrow{b}_i)\\
\overleftarrow{f} & = & \sigma(\overleftarrow{W}_{f}\cdot m_t + \overleftarrow{U}_{f}\cdot \overleftarrow{h}_{t+1} + \overleftarrow{b}_f)\\
\overleftarrow{o} & = & \sigma(\overleftarrow{W}_{o}\cdot m_t + \overleftarrow{U}_{o}\cdot \overleftarrow{h}_{t+1} + \overleftarrow{b}_o)\\
\overleftarrow{z} & = & \tanh(\overleftarrow{W}_{z}\cdot m_t + \overleftarrow{U}_{z}\cdot \overleftarrow{h}_{t+1} + \overleftarrow{b}_z)\\
\end{array}
$$
As is often done with bidirectional LSTM chains, all weights of the parameter matrices in the forward chain $\overrightarrow{W}, \overrightarrow{U}, \overrightarrow{b}$ are shared among all cells of the forward chain, and all weights of the parameter matrices in the backward chain $\overleftarrow{W}, \overleftarrow{U}, \overleftarrow{b}$ are shared among all cells of the backward chain. As in \citet{Molla2017}'s framework, there are separate sets of parameter matrices for the sentence and for the question.
\end{description}

\citet{Molla2017} used all the sentences of the full abstracts as the candidate input. Given that subsequent experiments observed an improvement of results by using the snippets only, our entries to BioASQ~6b used the snippets only. Also, \citet{Molla2017} observed very competitive results of a simple baseline that returned the first $n$ sentences. This suggests that sentence position is a useful feature and we consequently incorporated the sentence position as a feature in stage~3.

Given the sentence embedding $s_i$ and question embedding $q_i$, each obtained either by the mean of embeddings or by applying LSTM chains as described above, and given the sentence position $p_i$, stage~3 is implemented as a simple neural network with one hidden layer with a relu activation, followed by a linear combination:

$$
\begin{array}{rcl}
r_i & = & \max(0, W_r\left[s_i;q_i\odot s_i;p_i\right]+b_r)\\
score_i & = & W_{score}r_i + b_{score}\\
\end{array}
$$

Following \citet{Molla2017}'s framework, we used the element-wise multiplication between $q_i$ and $s_i$ as a way to encode the similarity between the question and the input sentence.

Figure~\ref{fig:results-comparison} compares the results of the deep learning approaches against the following baselines:
\begin{figure}
\centering
\includegraphics[width=\linewidth]{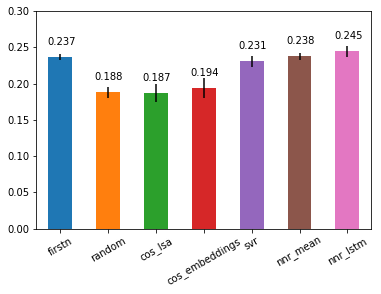}
\caption{Comparison of deep learning experiments with several baselines. The error bars indicate the standard deviation of 10-fold cross-validation.\label{fig:results-comparison}}
\end{figure}
\begin{description}
\item[Firstn.] Return the first $n$ sentences. As mentioned above, this baseline is often rather hard to beat.
\item[Random.] Return $n$ random sentences. This is the lower bound of any extractive summarisation system.
\item[Cos LSA.] Return the $n$ sentences that have the highest cosine similarity with the question. The feature vectors for the computation of the cosine similarity were obtained by computing $tf.idf$, followed by a dimension reduction step that selected the top 200 components after Latent Semantic Analysis.
\item[Cos Embeddings.] Return the $n$ sentences that have the highest cosine similarity with the question. The cosine similarity was based on the sum of the word embeddings (using embeddings with 200 dimensions) in the sentence/question.
\item[SVR.] Train a Support Vector Regression system that uses as features a combination of $tf.idf$, word embeddings and distance metrics as described by \citet{Molla2017}, plus the position of the snippet.
\end{description}

Figure~\ref{fig:results-comparison} shows that the "firstn" baseline is indeed hard to beat, and was matched only by the deep learning frameworks. Of these, the system using LSTM obtained the best results and was chosen for submission to BioASQ.

\section{Reinforcement Learning}\label{sec:rl}

\begin{figure}
\centering
\includegraphics[width=\linewidth]{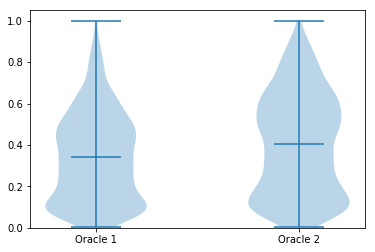}
\caption{Violin plots of the F1 ROUGE-SU4 scores of two oracles\label{fig:rouge-rouge-2}.}
\end{figure}

While the results using deep learning are encouraging, the models are trained on individually annotated sentences, and a summary is obtained by selecting the top $k$ sentences. An upper bound of the results obtained using such an approach would be an oracle that selects the $k$ sentences with highest individual ROUGE scores. Figures~\ref{fig:rouge-rouge-1} and~\ref{fig:rouge-rouge-2} show a comparison between the following two oracles:
\begin{figure}
\centering
\includegraphics[width=\linewidth]{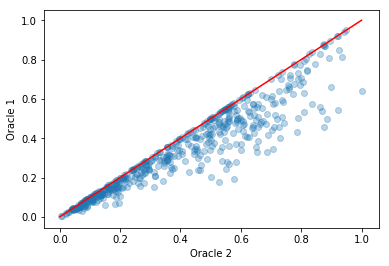}
\caption{Scatter-plot comparing the F1 ROUGE-SU4 score of two oracles using a random sample of 500 questions\label{fig:rouge-rouge-1}.}
\end{figure}
\begin{description}
\item[Oracle 1.] Return the $k$ snippets with highest \textit{individual} ROUGE score. This is a reasonable upper bound of supervised machine learning approaches such as presented in Section~\ref{sec:deep}.
\item[Oracle 2.] Return the combination of $k$ snippets with highest \textit{collective} ROUGE score. In particular, the oracle calculates the ROUGE score of every combination of $k$ snippets, and selects the combination with highest ROUGE score. This is an upper bound of any conceivable extractive summarisation system that returns $k$ snippets.
\end{description}

Figure~\ref{fig:rouge-rouge-1} shows the scatter-plot between oracles~1 and~2. It shows that oracle~1 under-performs oracle 2 in a number of questions. Figure~\ref{fig:rouge-rouge-2} plots the distributions of the ROUGE scores of each oracle side by side, and it clearly shows that the mean of the ROUGE scores of oracle~1 is lower than that of oracle~2.

Reinforcement learning allows to train the system on the ROUGE score of the final summary. This is done by iteratively allowing the system to extract a summary based on its current policy, and then updating the policy based on the feedback given by the ROUGE score of the extracted summary.

The reinforcement learning approach in our system is based on~\citet{Molla:alta2017}'s method. In particular, the reinforcement learning agent receives as input a candidate sentence and additional context information, and uses a global policy to determine the best possible action (either to select the sentence or not to select it). The global policy is implemented as a neural network with a hidden layer and is trained on a training partition of the development data by applying a variant of REINFORCE \cite{Williams1992}. 

More specifically, the global policy learns a set of parameters $\theta$ so that the policy predicts the probability of not selecting the sentence ($Pr(a=0;\theta)$) by applying the following neural network:
$$
\begin{array}{rcl}
Pr(a=0;\theta) & = & \sigma(W_hh + b_h)\\
h & = & \max(0, W_ss + b_s)  
\end{array}
$$
where $\theta = \left[W_h;W_s;b_h;b_s\right]$ and the input $s$ is the concatenation of the following features:
\begin{enumerate}
\item \emph{tf.idf} of candidate sentence $i$;
\item \emph{tf.idf} of the entire input text to summarise;
\item \emph{tf.idf} of the summary generated so far;
\item \emph{tf.idf} of the candidate sentences that are yet to be
  processed;
\item \emph{tf.idf} of the question; and
\item Length (in number of sentences) of the summary generated so far.
\end{enumerate}

The features chosen are such that the global policy has information about the candidate sentence (1), the entire list of candidate sentences (2), the summary that has been generated so far (3), the input sentences that are yet to be processed (4), and the question (5). Experiments by~\citet{Molla:alta2017} show that this information suffices to learn a global policy. In addition, we added the length of the summary generated so far (6). Our preliminary experiments showed that this additional feature facilitated a faster learning of the policy, and produced better results overall.

The specific algorithm that learns the global policy is presented in Figure~\ref{fig:policy}. 
\begin{algorithm}
    \KwData{$\mathtt{train\_data}$}
    \KwResult{$\theta$}
    $\mathtt{sample} \sim Uniform(\mathtt{train\_data})$\;\label{alg:sample}
    $s \leftarrow \mathtt{env.reset}(\mathtt{sample})$\;
    $\mathtt{all\_gradients} \leftarrow \emptyset$\;
    $\mathtt{initialise}(\theta)$\;\label{alg:init}
    $\mathtt{episode} \leftarrow 0$\;\label{alg:episode1}
    \While{True}{
    $\xi \sim Bernoulli\left(\frac{Pr(a=0;\theta)+p}{1+2\times p}\right)$\;\label{alg:sample2}
    $y \leftarrow 1-\xi$\;
    $\mathtt{gradient} \leftarrow \frac{\nabla(\hbox{cross\_entropy}(y,Pr(a=0;\theta))}{\nabla \theta}$\;
    $\mathtt{all\_gradients.append}(\mathtt{gradient})$\;
    $s, r, done \leftarrow \mathtt{env.step}(\xi)$\;\label{alg:reward}
    $\mathtt{episode} \leftarrow \mathtt{episode} + 1$\;\label{alg:episode2}
    \If{done}{\label{alg:done}
    $\theta \leftarrow \theta - \alpha\times r \times \hbox{mean}(\mathtt{all\_gradients})$\;\label{alg:apply}
    $\mathtt{sample} \sim Uniform(\mathtt{train\_data})$\;\label{alg:sample3}
    $s \leftarrow \mathtt{env.reset}(\mathtt{sample})$\;
    $\mathtt{all\_gradients} \leftarrow \emptyset$\;
    }
    }
  \caption{Training by Policy Gradient, where
    $\theta = \left[W_h;W_s;b_h;b_s\right]$.} 
  \label{fig:policy}
\end{algorithm}
The system first chooses one question from the training data (line~\ref{alg:sample}). Then, after randomly initialising the parameters of the global policy (line~\ref{alg:init}), it iteratively samples an action from the current global policy plus some perturbation $p$ (line~\ref{alg:sample2}) and applies the action (line~\ref{alg:reward}). When all the candidate sentences related to the question have been processed and actioned on (line~\ref{alg:done}), the resulting summary is evaluated and a reward $r$ produced (done previously, in line~\ref{alg:reward}). Then, the policy parameters are updated by multiplying the mean of all gradients obtained at every step by the reward (line~\ref{alg:apply}), and a new question is selected from the training data (line~\ref{alg:sample3}). Each iteration step is called an episode (lines~\ref{alg:episode1} and~\ref{alg:episode2}).

The perturbation~$p$ forces a wide exploration of the action space during the first episodes and is gradually reduced at every episode according to this formula:
$$
p = 0.2 \times 3000 / (3000 + \hbox{episode})
$$

\subsection{ROUGE Variants}

The evaluation scripts of the BioASQ task used the original Perl implementation of ROUGE-2 and ROUGE-SU4 \cite{Lin:2004}. Our experiments aimed to use ROUGE-SU4. 
Whereas the Perl implementation of ROUGE was used for the deep learning experiments described in section~\ref{sec:deep}, as an implementation decision we used Python's \texttt{pyrouge} library for the reinforcement learning experiments. Python's \texttt{pyrouge} provides ROUGE-1, ROUGE-2 and ROUGE-L, but not ROUGE-SU4. 

\begin{figure}
\centering
\includegraphics[width=\linewidth]{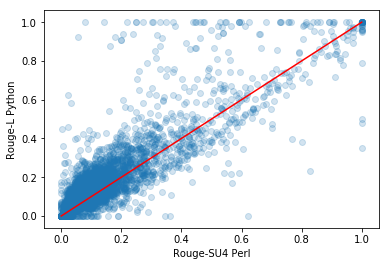}
\caption{Comparison between the Python and Perl implementations of ROUGE for a random sample of 5000 snippets and their respective ideal answers. The Python implementation uses ROUGE-L. The Perl implementation uses ROUGE-SU4.\label{fig:rouge-perl-python-1}}
\end{figure}

\begin{figure}
\centering
\includegraphics[width=\linewidth]{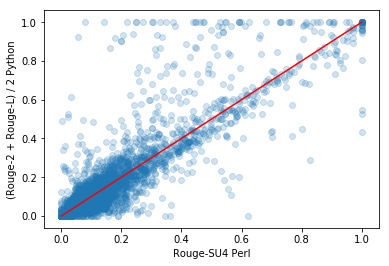}
\caption{Comparison between the Python and Perl implementations of ROUGE for a random sample of 5000 snippets and their respective ideal answers. The Python implementation uses (ROUGE-2 + ROUGE-L) / 2. The Perl implementation uses ROUGE-SU4.\label{fig:rouge-perl-python-2}}
\end{figure}

Figures~\ref{fig:rouge-perl-python-1} and~\ref{fig:rouge-perl-python-2} compare the ROUGE F1 scores of the Python libraries against the ROUGE-SU4 F1 score of the Perl implementation.
Figure~\ref{fig:rouge-perl-python-1} uses the Python library for ROUGE-L, and Figure~\ref{fig:rouge-perl-python-2} uses the Python library for the mean between ROUGE-2 and ROUGE-L.  We can observe some noise in the correlation between the Perl and Python implementations, but in general the mean between ROUGE-2 and ROUGE-L was a better approximation of the Perl implementation of ROUGE-SU4.

In general, we observed lower results of the Python versions of ROUGE-L and ROUGE-2 compared with the Perl versions. On the light of this, we strongly advise always to specify the implementation of ROUGE being used, since the results produced by different versions may not be comparable.

\section{Settings}\label{sec:settings}

The snippets were split into sentences using NLTK's sentence tokeniser.

The $tf.idf$ information used for the baselines was computed using NLTK's TfidfVectorizer. As in the system by \citet{Molla2017}, this vectoriser was trained using a collection of text consisting of the questions and the ideal answers.

The SVR experiments were implemented using Python's \texttt{sklearn} library and used word embeddings with 200 dimensions. The specific settings of the SVR model are shown in Table~\ref{tab:settings-svr}.
\begin{table}
\centering
\begin{tabular}{lccc}
Kernel & C & gamma \\
\midrule
rbf & 1.0 & 0.1 \\
\end{tabular}
\caption{Settings used in the SVR experiments.\label{tab:settings-svr}}
\end{table}

The deep learning experiments were implemented using TensorFlow's libraries. The specific details of the model are:
\begin{itemize}
\item Dimension of embeddings: 100
\item Length of the LSTM chain: 300. Sentences with more than 300 words were truncated.
\item Dimension of $\overrightarrow{h}$ and $\overleftarrow{h}$: 100 
\item Number of cells in the final hidden layer: 50
\end{itemize}

Table~\ref{tab:settings-nnr} shows the training settings used by the deep learning architectures. These settings were obtained empirically in a fine-tuning stage.
\begin{table}
\centering
\begin{tabular}{lccc}
System & Batch & Dropout & Epochs \\
\midrule
Mean & 4096 & 0.4 & 5\\
LSTM & 4096 & 0.8 & 10\\
\end{tabular}
\caption{Settings used in the deep learning experiments.\label{tab:settings-nnr}}
\end{table}

The reinforcement learning experiments were implemented in TensorFlow. Due to hardware constraints, the reinforcement learning approach only processed the first 20 sentences. The specific details of the architecture of the global policy are:
\begin{itemize}
\item Number of cells in the hidden layer: 200
\end{itemize}

The global policy was trained using a training partition and evaluated on a separate partition. The global policy parameters that generated the best results in the evaluation partition were used for the final runs to BioASQ.

\section{Results}\label{sec:results}

We submitted 5 runs for each batch as listed below.

\newcommand{\runname}{MQ-}

\begin{description}
\item[\runname1:] Return the first $n$ sentences. This is the Firstn baseline described in Section~\ref{sec:deep}. 
\item[\runname2:] Return the $n$ sentences with highest cosine similarity with the question. This is the Cos Embeddings baseline described in Section~\ref{sec:deep}.
\item[\runname3:] Return the $n$ sentences according to the SVR baseline described in Section~\ref{sec:deep}.
\item[\runname4:] Score the sentences using the LSTM-based deep learning approach as described in Section~\ref{sec:deep}.
\item[\runname5:] Apply reinforcement learning as described in Section~\ref{sec:rl}, with the variations described below.
\end{description}

The policy trained for the reinforcement learning approach of run \runname5 varied in several of the batches. In particular, the first batch was trained on ROUGE-L while batches~2 to~5 were trained on (ROUGE-2 + ROUGE-L) / 2. Also, to test the impact of different initialisation settings, we trained the system twice and generated two separate models. Batch~2 used one model, whereas batches~3 to~5 used the other model. 

\begin{figure}
\centering
\includegraphics[width=\linewidth]{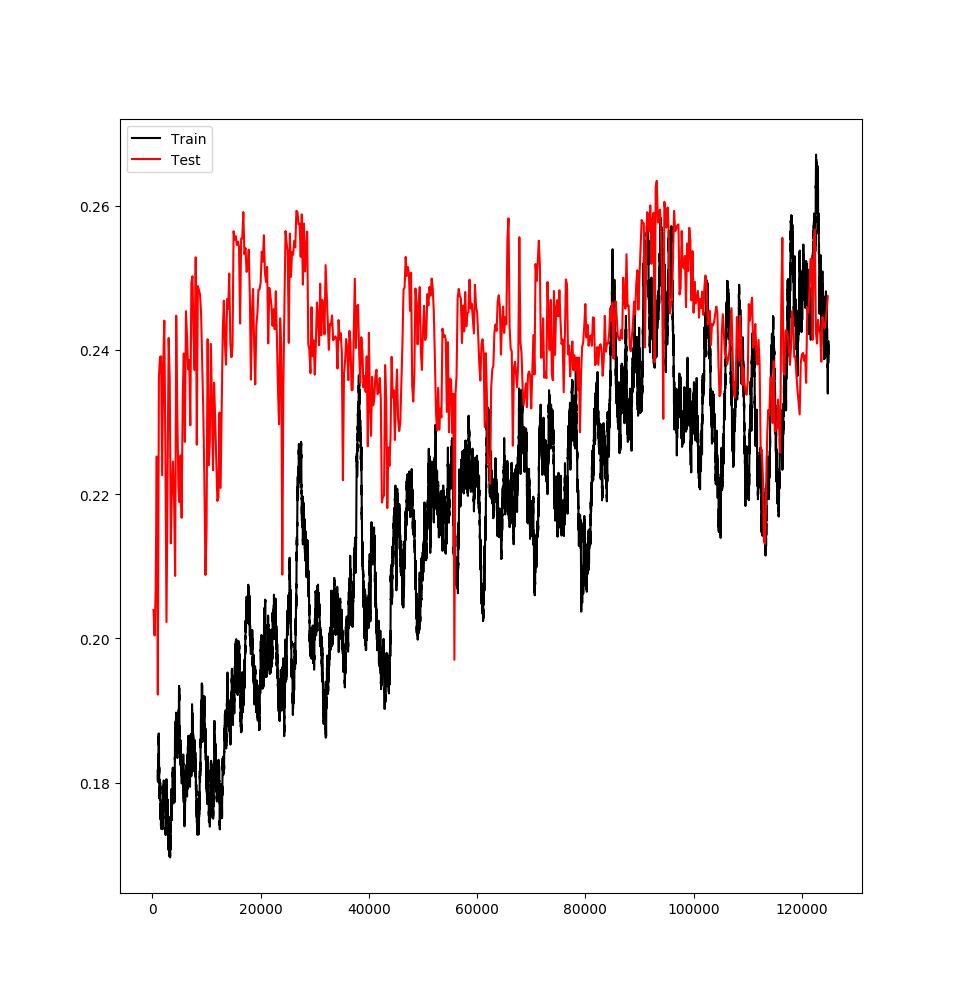}
\caption{Reinforcement Learning model trained for batch 1. The reward (y axis) is ROUGE-L.\label{fig:rl-1}}
\end{figure}

\begin{figure}
\centering
\includegraphics[width=\linewidth]{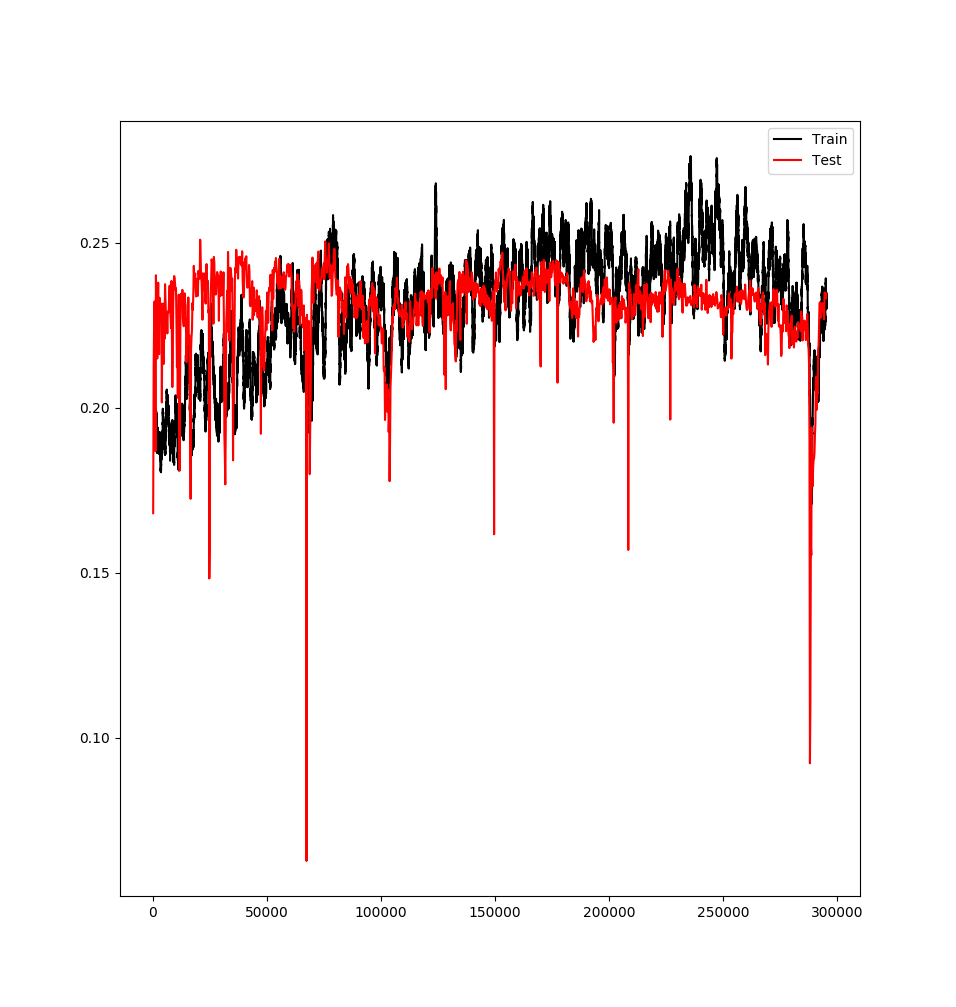}
\caption{Reinforcement Learning model trained for batch 2. The reward (y axis) is (ROUGE-2+ROUGE-L)/2.\label{fig:rl-2}}
\end{figure}

Figures~\ref{fig:rl-1} and~\ref{fig:rl-2} 
show the evolution of the results of the policies during the training stage for batches 1 and 2.
We observe some differences during training, but in general the best model on the test set achieved a ROUGE score between~0.25 and~0.26.\footnote{The training stage for batches 3 to 5 achieved a best result slightly over 0.26.} This is higher than the results reported by \citet{Molla:alta2017}, who reported a ROUGE score of about~0.2. The likely cause of the improvement in the results is the inclusion of the length of the summary generated so far in the context available by the policy.

\begin{figure}
\centering
\includegraphics[width=\linewidth]{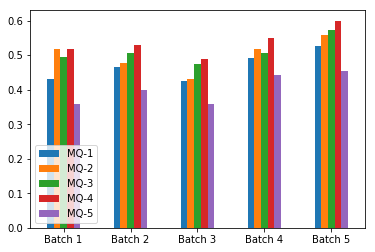}
\caption{ROUGE-SU4 results of the BioASQ runs.\label{fig:bioasqresults}}
\end{figure}

Figure~\ref{fig:bioasqresults} shows the results of the submissions to BioASQ.
In general, the deep learning approach (\runname4) achieved the best results. While the "first~n" baseline (\runname1) was fairly competitive and outperformed some of the runs of other participants to BioASQ, the baseline was not as strong as reported by \citet{Molla2017} on BioASQ5b.

We can also observe that the evaluation results of the submissions to BioASQ give higher ROUGE scores than those obtained in our experiments, both using the original Perl implementation of ROUGE, and the Python version. In fact, all of the runs except for \runname5 reported better BioASQ results than our experiments with the oracles. This is worth investigating.

The runs using reinforcement learning (\runname5) gave worse results than the other runs. The cause for this is also worth investigating, especially considering that, in our experiments, the results of the reinforcement learning approach were very competitive compared with the results of the other approaches.

\section{Conclusions}\label{sec:conclusions}

In this paper we have described the deep learning and reinforcement learning approaches used for the runs submitted to BioASQ 6b, phase B, for the generation of ideal answers. 

The deep learning approach used a supervised regression set-up to score the individual candidate sentences. The training data was generated by computing the ROUGE score of each individual candidate sentence, and the summary was obtained by selecting the top-scoring sentences. The results of the deep learning runs outperformed all other runs.

The reinforcement learning approach trained a global policy using as a reward the ROUGE score of the summary generated by the policy. The results of our experiments were very competitive but the submission results were lower than those of the other runs.

Further work will focus on the refinement of the reinforcement learning approach. In particular, further work will include the addition of a baseline in the training of the policy, as it has been shown to reduce the variance and to speed up the training process. Also, the architecture of the neural network implementation of the policy will be revised and enhanced by incorporating a more sophisticated model.

Further work on the deep learning runs will focus on the incorporation of more complex models. For example, preliminary experiments seem to indicate that a classification-based approach could outperform the current regression-based approach. Also, it is expected that a sequence-labelling approach would produce better results since the candidate sentences would not be processed independently.

\section*{Acknowledgments}
%

This research is partly funded by CSIRO's Data61.

\bibliography{bioasq2018}
\bibliographystyle{acl_natbib_nourl}

\end{document}